# Collaborative Optimization in Financial Data Mining Through Deep Learning and ResNeXt

Pengbin Feng
University of Southern California
Los Angeles, USA

Yankaiqi Li*
University of Wisconsin–Madison
Wisconsin, USA

Yijiashun Qi
University of Michigan
Ann arbor, USA

Xiaojun Guo
Independent Researcher
Jersey City, USA

Zhenghao Lin
Northeastern University
Boston, USA

*Abstract*—This study proposes a multi-task learning framework based on ResNeXt, aiming to solve the problem of feature extraction and task collaborative optimization in financial data mining. Financial data usually has the complex characteristics of high dimensionality, nonlinearity, and time series, and is accompanied by potential correlations between multiple tasks, making it difficult for traditional methods to meet the needs of data mining. This study introduces the ResNeXt model into the multi-task learning framework and makes full use of its group convolution mechanism to achieve efficient extraction of local patterns and global features of financial data. At the same time, through the design of task sharing layers and dedicated layers, it is established between multiple related tasks. Deep collaborative optimization relationships. Through flexible multi-task loss weight design, the model can effectively balance the learning needs of different tasks and improve overall performance. Experiments are conducted on a real S&P 500 financial data set, verifying the significant advantages of the proposed framework in classification and regression tasks. The results indicate that, when compared to other conventional deep learning models, the proposed method delivers superior performance in terms of accuracy, F1 score, root mean square error, and other metrics, highlighting its outstanding effectiveness and robustness in handling complex financial data. This research provides an efficient and adaptable solution for financial data mining, and at the same time opens up a new research direction for the combination of multi-task learning and deep learning, which has important theoretical significance and practical application value.

## I. Introduction

In recent years, the rapid advancement of artificial intelligence has established deep learning as a pivotal tool in data mining. Financial data, characterized by high dimensionality, heterogeneity, and time-series nature, often challenges traditional data mining approaches. Deep learning, particularly Convolutional Neural Networks (CNNs) [1], offers robust solutions for extracting features from complex and unstructured financial data. Beyond their established role in image analysis, CNNs, especially ResNeXt architectures, have demonstrated efficacy in time-series and structured data mining, effectively capturing local features and processing high-dimensional data through group convolution [2].

In financial data mining, multi-task learning (MTL) addresses the limitations of single-task models by leveraging shared feature spaces to enhance learning efficiency and generalization [3]. Integrating ResNeXt into MTL frameworks enables the exploration of spatiotemporal characteristics and establishes deep inter-task associations, facilitating tasks like trend prediction [4], risk assessment, and asset pricing [5]. This shared representation design enhances data utilization and model performance. Compared to traditional feature engineering, the ResNeXt-based MTL framework offers notable advantages. Its group convolution mechanism reduces parameter complexity while maintaining robust modeling capabilities, and efficiently handling high-dimensional data [6]. The framework's multi-task design allows flexible adjustment of task weights and loss functions, capturing global features while addressing specific task characteristics. This adaptability suits the multimodal nature of financial data, enabling higher-dimensional insights into market dynamics [7].

ResNeXt's residual block structure enhances gradient flow and feature reuse [8], addressing the volatility and complexity of financial data. Its grouping strategy preserves data diversity and information richness while capturing high-level semantic features, delivering comprehensive feature representations for financial tasks.

This study's ResNeXt-based MTL framework efficiently handles complex financial data and achieves collaborative optimization across tasks. Its innovative feature extraction and shared learning design provide a flexible, scalable solution for multi-task financial analysis. As financial data grows in scale and complexity, this framework offers robust support for

## II. RELATED WORK

In The integration of deep learning techniques with multi-task learning has enabled significant advancements in feature extraction, optimization, and prediction across complex datasets. The proposed framework leverages these advancements, particularly through the use of ResNeXt, to address challenges in financial data mining.

Graph-based methods have contributed significantly to understanding relationships and dependencies in structured data, which aligns with the goal of extracting correlations in financial datasets. Wei et al. explored self-supervised Graph Neural Networks (GNNs) for enhanced feature extraction, demonstrating their ability to capture intricate data patterns [9]. Similarly, Zhang et al. presented robust GNN methodologies to handle dynamic data structures, emphasizing their capability for maintaining stability across tasks [10]. These approaches inform the collaborative optimization strategies in multi-task learning.

Hybrid deep learning architectures, such as CNN-LSTM models, have shown substantial promise in handling temporal and sequential data, making them particularly relevant for capturing the time-series nature of financial data. Yao et al. utilized a hybrid CNN-LSTM model to improve prediction accuracy in sequence data, showcasing the architecture's capability for combining local and global pattern recognition [11]. Wang et al. expanded on this by incorporating spatiotemporal prediction techniques, offering insights into managing dynamic patterns effectively [12].

Self-supervised learning frameworks provide a robust pathway for overcoming challenges in data sparsity and feature generalization, which are critical in financial datasets. Xiao's work on self-supervised learning highlights how these methods can enhance representation learning in low-data scenarios, a feature applicable to diverse financial tasks [13]. Similarly, Shen et al. demonstrated semi-supervised learning's potential to improve feature extraction and predictive accuracy, reinforcing the adaptability of these techniques in data-constrained environments [14].

Transformer-based and encoder-decoder models have advanced contextual feature extraction, a vital component in analyzing multi-dimensional financial data. Chen et al. proposed a transformer-based framework for extracting coherent and high-quality features, which aligns with the extraction of high-dimensional patterns in financial data [15]. Additionally, Feng et al. combined data augmentation and few-shot learning strategies to tackle limited data scenarios effectively, a critical aspect of handling financial data with sparse labels [16].

Generative models and metric learning approaches have further enriched the ability to capture latent features in complex datasets. Liu et al. introduced calibration learning to improve the adaptation of few-shot learning models, providing insights into optimizing task-specific feature representation [17]. Luo et al. employed metric learning techniques to enhance model robustness under sparse data conditions, which is highly relevant for financial data modeling [18].

Techniques for large-scale feature extraction and adaptive modeling have also contributed significantly to understanding complex data patterns. Yang et al. investigated methods for adaptive feature extraction, which parallels the multi-dimensional feature modeling required in financial data analysis [19]. Additionally, Sun et al. optimized convolutional neural networks to improve pattern recognition, providing transferable techniques for enhancing deep feature extraction [20].

Graph-based techniques for reasoning and relationship modeling have also informed the shared learning component of the proposed framework. Du et al. demonstrated the utility of GNNs for entity extraction and reasoning, highlighting their relevance for capturing task relationships in shared spaces [21]. These methodologies directly contribute to the collaborative optimization mechanism central to this work.

## III. METHOD

This study proposes a multi-task learning framework based on ResNeXt, which aims to solve the problem of feature sharing and collaborative modeling between multiple tasks in financial data mining. The overall model is divided into three parts: feature extraction module, shared learning module, and task-specific head. The feature extraction module is built on ResNeXt and can fully capture the local patterns and global features of financial data; the shared learning module enhances the feature interaction between tasks through multi-task loss design [22]; the task-specific head generates the final prediction results for different tasks [23]. The main architecture of ResNext is shown in Figure 1.

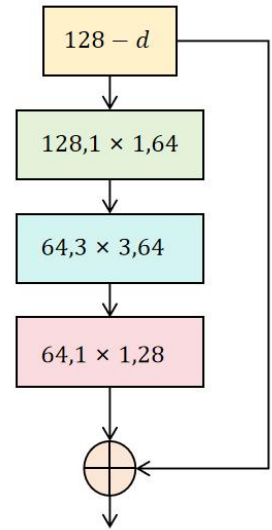

Figure 1 Overall architecture of the model

First, for the input financial data, it is represented in matrix form as $X \in R^{T \times F}$, where T represents the length of the time series and F represents the feature dimension. In the feature extraction module (Figure 1), '128-d' denotes the feature vector's dimensionality post-group convolution in ResNeXt.

This value, a tunable hyperparameter, balances the need for detailed feature representation against computational demands, ensuring efficient processing of high-dimensional financial data. The feature extraction module adopts the ResNeXt network and uses its group convolution design to effectively process high-dimensional financial data while enhancing the representation ability of features through residual connections. Each group convolution operation in ResNeXt can capture the feature patterns within different groups and finally aggregate the features in the channel dimension. After a series of nonlinear transformations of convolution, pooling, and activation functions, the extracted features are represented as $F_{shared} \in R^{T' \times C}$, where $T'$ is the time dimension after downsampling and C is the number of feature channels. This shared feature serves as the basic representation of all tasks and provides support for subsequent multi-task learning.

In the shared learning module, a task weight-sharing mechanism is introduced to map the shared features to the specific space of each task. For the i-th task, its dedicated feature is expressed as:

$$F_i = \sigma(W_i F_{shared} + b_i) \quad (1)$$

Where $W_i$ and $b_i$ are the learnable weights of task I, and $\sigma$ represents the activation function. Through this mapping, the model can learn specific representations for each task based on shared features, avoiding interference between tasks while maintaining efficient feature sharing.

The design of the task-specific head uses a fully connected network to map task-specific features to the final output space. For classification tasks, its output is a probability distribution $y'_i$, calculated as:

$$y'_i = \text{softmax}(W_{mlp,i} F_i + b_{mlp,i}) \quad (2)$$

For the regression task, the output is a scalar $y''_i$, and the calculation formula is:

$$y''_i = W_{mlp,i} F_i + b_{mlp,i} \quad (3)$$

The above-mentioned dedicated head can be flexibly adjusted according to the needs of different tasks. For example, the classification task uses the cross entropy loss function, while the regression task uses the mean square error (MSE) loss function [24]. In order to achieve joint optimization of multiple tasks, the overall loss function is defined as the weighted sum of the losses of each task:

$$L = \sum_{i=1}^{N} \alpha_i L_i \quad (4)$$

Where N is the number of tasks, and $\alpha_i$ is the weight of the i-th task, which can be flexibly adjusted according to the importance of the task to ensure the balance between different tasks.

The model training adopts a phased optimization strategy. First, single-task pre-training is used to ensure that the dedicated head of each task can converge effectively; then, all tasks are jointly optimized in the multi-task framework to further improve the overall performance of the model. In order to accelerate convergence and improve the generalization ability of the model, data enhancement strategies are introduced during the training process, such as random cropping of time series data, smooth noise addition, etc., and the Adam optimizer and dynamic learning rate adjustment strategy are used.

The proposed multi-task learning framework based on ResNeXt not only utilizes the efficient feature extraction capability of ResNeXt but also realizes the collaborative optimization between tasks through the shared learning mechanism, providing an efficient and flexible solution for multi-task mining of financial data. The design of the model achieves a good balance between feature extraction, sharing and task optimization, providing technical support for dealing with complex financial data.

IV. EXPERIMENT

*A. Datasets*

This study selected the S&P 500 stock market dataset as the data source for the experiment. This dataset is a widely used public dataset in the field of financial data mining, which contains historical trading data and market indicators of all constituent stocks in the S&P 500 index. The dataset records daily trading data since 2000, including key information such as opening price, closing price, highest price, lowest price, and trading volume. It also provides macroeconomic indicators (such as GDP growth rate, unemployment rate, interest rate, etc.) as additional features. These data can fully reflect the dynamic characteristics of the financial market and provide a reliable basis for tasks such as trend prediction and risk assessment.

In the data preprocessing stage, according to the time series characteristics of stock trading data, the data is divided into fixed-length time windows, each of which contains several days of trading data to capture short-term market trends. At the same time, the macroeconomic indicators in the data are aligned to the corresponding time windows to supplement the global market background information. In addition, considering the common missing values and noise problems in financial data, the study interpolated the missing data and standardized the features through Z-Score normalization to eliminate the influence of feature dimensions on model training. These steps, when implemented with linked data [25], ensure the integrity and consistency of the data and provide high-quality input data for the model.

The dataset was finally divided into training set, validation set and test set. After cleaning, approximately 1.8 million samples remained for training (70%), validation (15%), and testing (15%). Preprocessing involved fixed-length time windows, interpolation of missing values, and Z-score normalization. The training set is used to optimize the model parameters, the validation set is used to adjust the hyperparameters and evaluate the generalization performance of the model, and the test set is used for the final model performance evaluation. In order to enhance the robustness of the model, this study introduced data enhancement strategies, such as random translation or scaling of time series data to increase the diversity of the data. These designs ensure that the use of the dataset can fully reflect the complexity of the financial market and provide a reliable basis for the verification of the multi-task learning model.

## B. Experiments

In order to comprehensively evaluate the performance of the proposed ResNeXt-based multi-task learning framework in financial data mining, this study selected four classic deep learning models as comparison objects: Long Short-Term Memory (LSTM), as the mainstream model for processing time series data, is good at capturing the temporal dependency of financial data; Transformer model, due to its powerful global attention mechanism, can effectively model the long-range dependency in financial data; Deep Shared Network (DSN) based on multi-task learning, through the design of shared layers and task-specific layers, it improves the collaborative optimization ability between tasks; and Multi-Channel Convolutional Neural Network (MCCNN), which uses multi-channel design to capture the correlation between different features and shows the ability to efficiently model multi-dimensional financial data. These models cover different strategies of sequence modeling, multi-task learning and multi-dimensional feature extraction, providing a sufficient comparison basis for verifying the superiority of the proposed method. The comparative experiments of the discriminant task and regression task of the multi-task model are shown in Tables 1 and 2.

Table 1 Results of the discrimination task experiment

| Model | Acc | Macro F1 |
|---|---|---|
| LSTM | 73.2 | 71.8 |
| DSN | 76.4 | 74.2 |
| MCCNN | 79.6 | 77.8 |
| Transformer | 82.1 | 80.4 |
| ours | 85.3 | 83.7 |

Table 2 Experimental results of regression task

| Model | MAE | RMSE |
|---|---|---|
| LSTM | 0.041 | 0.063 |
| DSN | 0.036 | 0.057 |
| MCCNN | 0.033 | 0.052 |
| Transformer | 0.029 | 0.048 |
| ours | 0.025 | 0.043 |

The proposed ResNeXt-based multi-task learning framework outperforms other models in both classification and regression tasks. In classification, it achieved an accuracy of 85.3% and a macro F1 score of 83.7%, significantly higher than the second-ranked multi-channel convolutional neural network (MCCNN). In regression, the mean absolute error (MAE) was 0.025, and the root mean square error (RMSE) was 0.043, demonstrating excellent prediction capabilities. These results highlight the framework's strength in processing complex financial data with enhanced feature extraction and task collaboration.

The classification task reveals the differences in model performance based on financial data characteristics. LSTM, despite its temporal dependence modeling capabilities, has lower accuracy and macro-average F1 scores compared to other models. Transformer improves time series modeling and long-range dependencies but lacks the feature extraction module to fully utilize financial data diversity and local patterns, resulting in lower classification performance.

The DSN model in the multi-task learning framework further improved its performance in classification tasks, with an accuracy rate of 79.6% and a macro-average F1 of 77.8%. This result shows that through the design of shared and dedicated layers, DSN can more effectively capture the potential correlations between different tasks, thereby improving the classification performance of the model. However, the feature extraction module of DSN is relatively simple and fails to fully mine the complex patterns in financial data, so its performance still lags behind MCCNN and proposed methods. MCCNN captures the relationship between different dimensions of financial features through multi-channel design, and the classification performance is significantly improved, with an accuracy of 82.1%. However, its lack of in-depth modeling of time series and shared features between tasks limits further improvement.

In regression tasks, the advantages of the proposed method are particularly obvious. In comparison to other models, the proposed framework not only effectively captures local patterns and global features of financial data using ResNeXt, but also facilitates deep collaborative optimization between tasks through a multi-task learning mechanism, resulting in the lowest values for both MAE and RMSE. In contrast, although LSTM and Transformer have certain advantages in modeling time series, they lack an effective feature sharing mechanism, resulting in limited performance in multi-task environments. In addition, although MCCNN has made breakthroughs in multi-dimensional feature modeling, its regression performance is not as good as the proposed method due to its insufficient ability to capture time dependence and global features.

Overall, the experimental results fully illustrate the wide applicability and excellent performance of the proposed ResNeXt-based multi-task learning framework in financial data mining. Through an efficient feature extraction mechanism and shared learning design, the model can not only achieve accurate predictions in classification tasks but also effectively reduce errors in regression tasks. These results verify the potential of ResNeXt combined with multi-task learning and also demonstrate its powerful ability to process complex and diverse financial data. In the future, this framework can be further extended to other financial scenarios, such as risk management and market monitoring, to help the development of intelligent financial analysis.

## V. CONCLUSION

This study proposes a multi-task learning framework based on ResNeXt and conducts an in-depth exploration of feature extraction and multi-task collaborative optimization issues in financial data mining. Experimental results show that this framework can achieve significantly better performance than other models in classification and regression tasks, especially when processing complex time series features and multi-dimensional data patterns, showing powerful modeling capabilities. By combining ResNeXt's group convolution mechanism and the shared optimization design of multi-task learning, the model achieved optimal performance in terms of accuracy, F1 score, and error indicators, verifying the effectiveness and reliability of the method.

In addition, the framework is designed with good scalability and applicability, and can flexibly deal with high-dimensional features, nonlinear relationships, and noise problems in financial data. It improves the efficiency of multi-task learning through feature sharing between tasks and also enhances the model's ability to capture complex patterns in financial data. This research not only provides an efficient solution for financial data mining but also provides a strong reference for the application of multi-task learning frameworks in other fields, demonstrating the broad prospects of deep learning in complex data scenarios.

Future research can expand the capabilities of this framework in multiple directions. On the one hand, you can explore application scenarios for larger-scale financial data, such as adding real-time high-frequency data to enhance the dynamic prediction capabilities of the model; on the other hand, you can try to combine advanced models such as Graph Neural Network to better Model the complex correlations between different assets in financial markets. In addition, while optimizing the model, it will also be an important direction to study how to reduce computational complexity and improve operating efficiency so that it can operate in resource-constrained environments. These efforts will further promote the application of intelligent financial analysis and inject new impetus into the development of financial technology.